\documentclass[letterpaper, 10pt, conference]{ieeeconf}

\IEEEoverridecommandlockouts
\usepackage{cite}
\usepackage{amsmath,amssymb,amsfonts}
\usepackage{algorithmic}
\usepackage{graphicx}
\usepackage{textcomp}
\usepackage[table]{xcolor}
\usepackage{multirow}
\usepackage{microtype}
\usepackage{float}
\usepackage[caption=false]{subfig}
\usepackage[english]{babel}

\title{\LARGE \bf
Cross-subject Muscle Fatigue Detection via Adversarial and Supervised Contrastive Learning with Inception-Attention Network
}

\author{Zitao Lin, Chang Zhu, and Wei Meng, \textit{Senior Member, IEEE}
\thanks{*This work was supported in part by the National Natural Science Foundation of China under Grant 52275029 and the Natural Science Foundation of Hubei Province under Grant 2025AFD639. (Corresponding authors: Wei Meng). Z.~Lin, C.~Zhu, and W.~Meng are with the School of Information Engineering, Wuhan University of Technology, Wuhan 430070, China (Email: linzt2004@whut.edu.cn, changzhu@whut.edu.cn, and weimeng@whut.edu.cn)}}%

\begin{document}
\maketitle
\thispagestyle{empty}
\pagestyle{empty}

\begin{abstract}
Muscle fatigue detection plays an important role in physical rehabilitation. Previous researches have demonstrated that {sEMG} offers superior sensitivity in detecting muscle fatigue compared to other biological signals. However, features extracted from sEMG may vary during dynamic contractions and across different subjects, which causes unstability in fatigue detection. To address these challenges, this research proposes a novel neural network comprising an Inception-attention module as a feature extractor, a fatigue classifier and a domain classifier equipped with a gradient reversal layer. The integrated domain classifier encourages the network to learn subject-invariant common fatigue features while minimizing subject-specific features. Furthermore, a supervised contrastive loss function is also employed to enhance the generalization capability of the model. Experimental results demonstrate that the proposed model achieved outstanding performance in three-class classification tasks, reaching 93.54\% accuracy, 92.69\% recall and 92.69\% F1-score, providing a robust solution for cross-subject muscle fatigue detection, offering significant guidance for rehabilitation training and assistance.
\end{abstract}

\begin{keywords}
Muscle fatigue, sEMG, Domain adversary, Supervised contrastive learning
\end{keywords}

\section{Introduction}\label{sec:Introduction}
Muscle fatigue is defined as the decline in the capacity of muscles to generate force or power during activities \cite{constantin2021molecular}. While muscle fatigue can be estimated through subjective scales such as the Borg CR-10 rating of perceived exertion (RPE) scale \cite{frasie2024borg-cr10} or quantified through maximum voluntary contraction (MVC) \cite{zhang2024multilevel} or muscle oxygen saturation \cite{villafaina2023behavior}, detecting muscle fatigue using these objective metrics is often impractical for robust monitoring due to high physical demand and operational complexity. In contrast, sEMG has proven to be a highly reliable metric compared with other techniques \cite{Li2024non-invasive}. Numerous studies have demonstrated that sEMG features in the time and frequency domains, such as root mean square, median frequency and zero crossing rate, alternate during fatigue progression \cite{Wu2025time_and_freq_features} \cite{sun2025detecting}. However, these traditional indices remain unstable. Liu et al. observed that sEMG signals exhibit strong nonlinearity and non-stationarity, which leads to substantial uncertainties in time-domain and frequency-domain \cite{liu2023dynamic}. Similarly, Rampichini et al. highlighted the complex, non-linear characteristic of sEMG in response to fatigue \cite{rampichini2020complexity}, while Zhang et al. concluded that evaluating muscle fatigue through time-frequency attributes may oversimplify the physiological phenomenon, resulting in imprecise outcomes \cite{zhang2024multilevel}. Consequently, more sophisticated analytical methods, particularly deep-learning (DL), have become the mainstream approach in muscle fatigue detection.

DL-based methods can automatically learn complex features from raw signals with less manual feature extraction, achieving superior performance in complex pattern recognition. Recently, DL-based methods have been extensively deployed for muscle fatigue detection. Zhang et al. developed MACNet to classify muscle fatigue states during handgrip MVC tests \cite{zhang2024multilevel}, but the model trained on subject-wise trials could not generalize to unseen subjects. Although Zhang et al. introduced Multi-dimensional Feature Fusion Network (MFFNet) integrating attention-based time and frequency domain modules to facilitate cross-subject fatigue detection, reaching an accuracy of 78.25\% \cite{zhang2021mffnet}, robustness on cross-subject scenario can still be further enhanced. Additionally, DL-based methods often rely on stringent experimental constraints, such as constant force output or subjects with similar physical conditions, otherwise fatigue-related manifestations may be overshadowed by irrelevant inter-subject variance, leading to degraded performance in cross-subject fatigue detection.

The main contribution of this study is summarized as follows:

(1) a novel network called Inception-Attention-Domain-Adversarial-Net (IADAN) is proposed, incorporating a feature extractor, a domain classifier with gradient reversal layer (GRL) and a fatigue classifier to extract multi-scale feature from the input data.

(2) Adversarial mechanism and supervised contrastive learning strategy is introduced to further enhance the discriminative manifestations within each fatigue state while minimizing fatigue-irrelevant variance.

The remainder of this paper is organized as follows: Section II describes the architecture of the proposed model. Section III details the experimental protocol and data preprocessing. Section IV presents the results, along with ablation studies and comparative analysis. Finally, Section V concludes the paper and discusses the limitation.

\section{Methods}

\subsection{Feature Extractor}
The feature extractor is designed to capture multi-scale features from the processed sEMG signals. Two dilated convolutional layers with ReLU activations followed by batch normalization and max pooling layers are first employed. Unlike standard convolutions, dilated convolutional kernels incorporate a dilation rate to expand the receptive field. This allows the network to capture global features rather than being distracted by local artifacts \cite{yu2015dilation}. The receptive field $R$ for a single kernel is calculated by:
\begin{align}
        R = K + (K - 1)(D - 1)
\end{align}
where $K$ denotes the kernel size and $D=2$ represents the dilation rate.

An Attention module is integrated to prioritize informative regions and ignore redundant ones across both the channel and spatial dimensions. The Attention module consists of a Channel Attention Block (CAB) and a Spatial Attention Block (SAB). The CAB first compresses the spatial dimensions using both average and max pooling operations to aggregate global context. Following this, the SAB captures inter-spatial relationships by applying a convolutional layer over the pooled features \cite{woo2018cbam}.

\begin{figure}[H]
    \centering
    \includegraphics[width=0.9\linewidth]{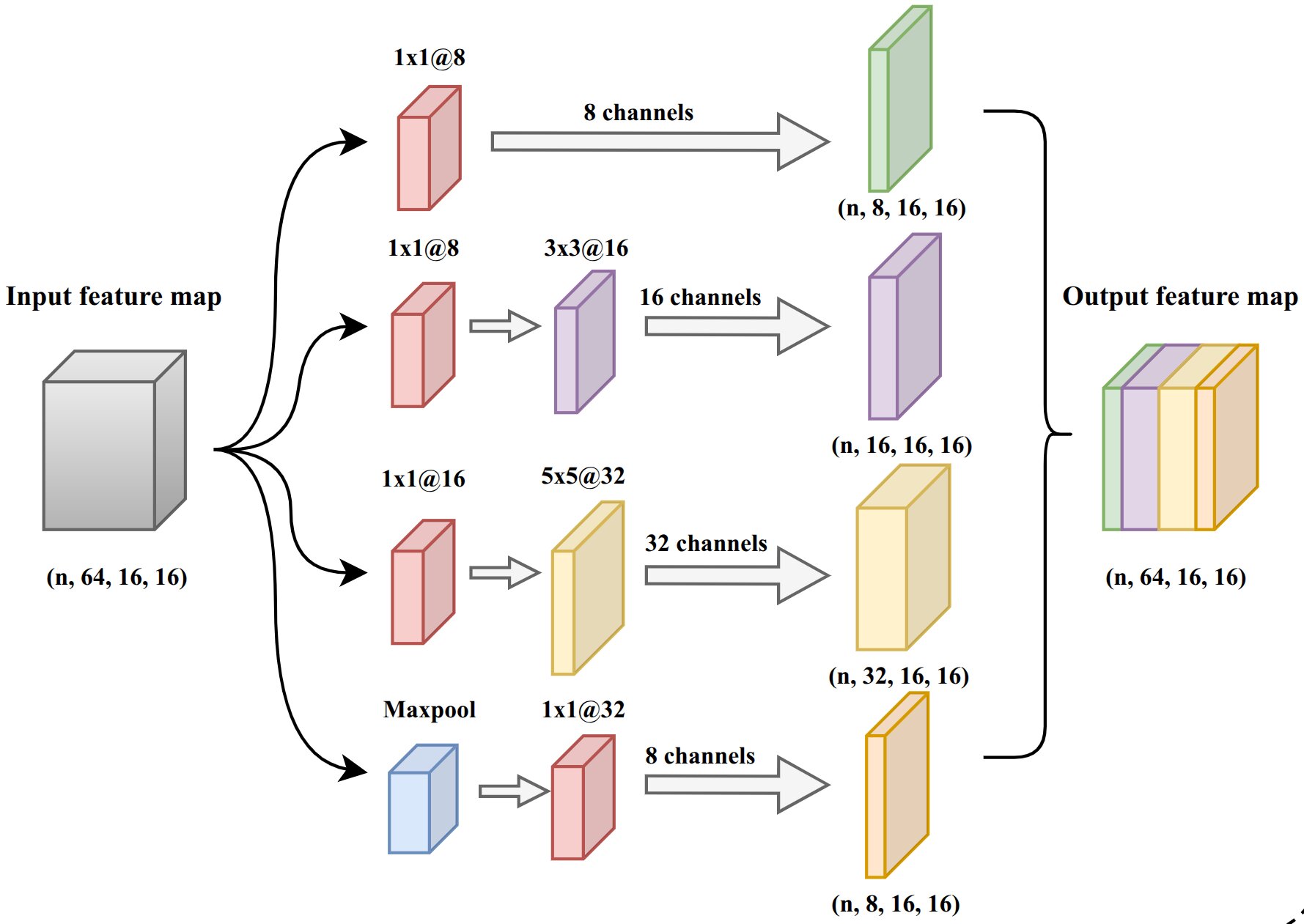}
    \caption{The structure of the Inception module.}
    \label{fig:Inception}
\end{figure}

Subsequently, an Inception module is employed to further enrich feature representation, which utilizes parallel multi-scale convolutional branches to capture features at different levels \cite{szegedy2016inception}. The $1 \times 1$ convolutional operation is first applied in each branch to reduce channel dimensionality, then the dimension-reduced feature map is processed through kernels of different sizes, as depicted in Fig.~\ref{fig:Inception}. The output of those branches are then concatenated along the channel dimension. 

\subsection{Adversarial and Supervised Contrastive Learning}

Domain adversarial training based on GRL is introduced for cross-subject generalization. During forward propagation, GRL functions as an identity operator, denoting $\text{GRL}(x)=x$. Conversely, during backpropagation, gradients are multiplied by a negative scalar defined as $-\lambda$ to reverse the gradient flow, denoting $\frac{d}{dx}\text{GRL}(x) = -\lambda$. This mechanism compels the feature extractor to minimize the domain classification accuracy, thereby suppressing subject-specific differences while preserving generalized fatigue-related features. $\lambda$ is dynamically adjusted to ensure stability during training, according to the following equation \cite{ganin2016domain}:
\begin{align}
    \lambda = k\left( \frac{2}{1 + e^{-\gamma p}} - 1 \right)
\end{align}
where $\gamma$ and $k$ are hyperparameters that control the growth rate of adversarial strength, empirically set to 5 and 0.2, respectively. $p$ is defined as the ratio of the current number of steps to the total number of steps.

the supervised contrastive (SupCon) loss function was employed To further enhance the generalization across different subjects. Unlike standard unsupervised learning, SupCon loss utilizes label information to treat all samples within a batch. Samples with the same fatigue label are categorized as positive pairs, while those with different fatigue labels are regarded as negative pairs~\cite{khosla2020supervised}. The loss $L_{SC}$ is given by:
\begin{align}\label{eq:SupCon}
    L_{SC} = \sum_{i \in I} \frac{-1}{|P(i)|} \sum_{p \in P(i)} \log \frac{\exp(z_i \cdot z_p / \tau)}{\sum_{a \in A(i)} \exp(z_i \cdot z_a / \tau)}
\end{align}
where $i \in I$ is the index of a sample in the batch. $P(i)$ denotes the set of indices of all positive samples. $A(i)$ is the set of all indices in the batch excluding $i$. $z$ represents the normalized feature vectors, and $\tau$ is a hyperparameter called temperature, controlling the concentration of the distribution, which is set to 0.05 as a typical value. 

\subsection{The Overall Structure}

\begin{figure*}[t]
    \centering
    \includegraphics[width=\textwidth]{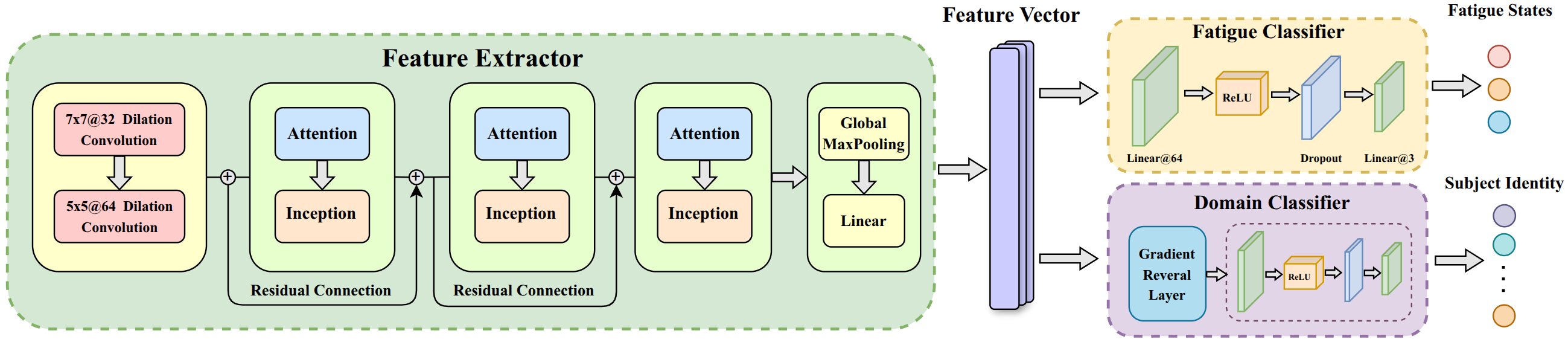}
    \caption{The overall structure of the proposed IADAN.}
    \label{fig:network structure}
\end{figure*}

\begin{table*}[htbp]
    \caption{The Model Structure Parameters} \label{tab:parameters}
    \centering
    \resizebox{0.96\linewidth}{!}{%
    \begin{tabular}{|c|c|c|c|c|c|}
    \hline
    \textbf{Modules} &
    \textbf{Layers} &
    \textbf{Kernel Size} &
    \textbf{\begin{tabular}[c]{@{}c@{}}Kernel/Neurons Numbers\end{tabular}} &
    \textbf{Input Shape} &
    \textbf{Output Shape} \\ \hline
    \multirow{7}{*}{\textbf{\begin{tabular}[c]{@{}c@{}}Feature\\ Extractor\end{tabular}}} &
    \textbf{Dilated Convolution 1} &
    7 & 32 & (n,6,64,64) & (n,32,32,32) \\ \cline{2-6} &
    \textbf{Dilated Convolution 2} &
    5 & 64 & (n,32,32,32) & (n,64,16,16) \\ \cline{2-6} &
    \textbf{Inception Branch 1} &
    1 & 8 & (n,64,16,16) & (n,8,16,16) \\ \cline{2-6} &
    \textbf{Inception Branch 2} &
    3 & 16 & (n,64,16,16) & (n,16,16,16) \\ \cline{2-6} &
    \textbf{Inception Branch 3} &
    5 & 32 & (n,64,16,16) & (n,32,16,16) \\ \cline{2-6} &
    \textbf{Inception Branch 4} &
    1 & 8 & (n,64,16,16) & (n,8,16,16) \\ \cline{2-6} &
    \textbf{Fully Connected Layer} &
    \multirow{5}{*}{\textbf{None}} & 128 & (n,64,16,16) & (n,128) \\ \cline{1-2} \cline{4-6}
    \multirow{2}{*}{\textbf{\begin{tabular}[c]{@{}c@{}}Fatigue\\ Classifier\end{tabular}}} &
    \textbf{Linear Layer 1} &
    & 64 & (n,128) & (n,64) \\ \cline{2-2} \cline{4-6}  &
    \textbf{Linear Layer 2} &
    & 3 & (n,64) & (n,3) \\ \cline{1-2} \cline{4-6} 
    \multirow{2}{*}{\textbf{\begin{tabular}[c]{@{}c@{}}Domain\\ Classifier\end{tabular}}} &
    \textbf{Linear Layer 1} &
    & 64 & (n,64) & (n,64) \\ \cline{2-2} \cline{4-6}  &
    \textbf{Linear Layer 2} &
    & 9 & (n,64) & (n,9) \\ \hline
    \end{tabular}%
    }
\end{table*}

IADAN integrates a feature extractor, followed by a fatigue classifier and a domain classifier. The overall structure is illustrated in Fig.~\ref{fig:network structure}, and the parameters for each layer are detailed in Table \ref{tab:parameters}. In the feature extractor, the attention and the Inception modules are alternately stacked three times to form the Inception-attention module, with residual connections to reduce the risk of degradation. The output is converted to a 128-dimensional vector through global max pooling and a fully-connected layer. This vector is simultaneously fed into the fatigue classifier for fatigue state recognition and the domain classifier for subject-wise generalization.

The fatigue classifier performs three-class fatigue state recognition tasks, consisting of two linear layers with ReLU activation and a dropout layer, whose output corresponds to the three fatigue levels, defined as non-fatigue (NF), medium-Fatigue (MF) and severe-fatigue (SF). The detailed definitions are described in Section \ref{sec:experiment}. The domain classifier consists of a GRL and a fully connected network structured similarly to the fatigue classifier but terminating in a different number of neurons in the output layer, corresponding to the number of subjects in the training dataset, with each subject being regarded as a domain. 

\begin{figure}[H]
    \centering
    \includegraphics[width=\linewidth]{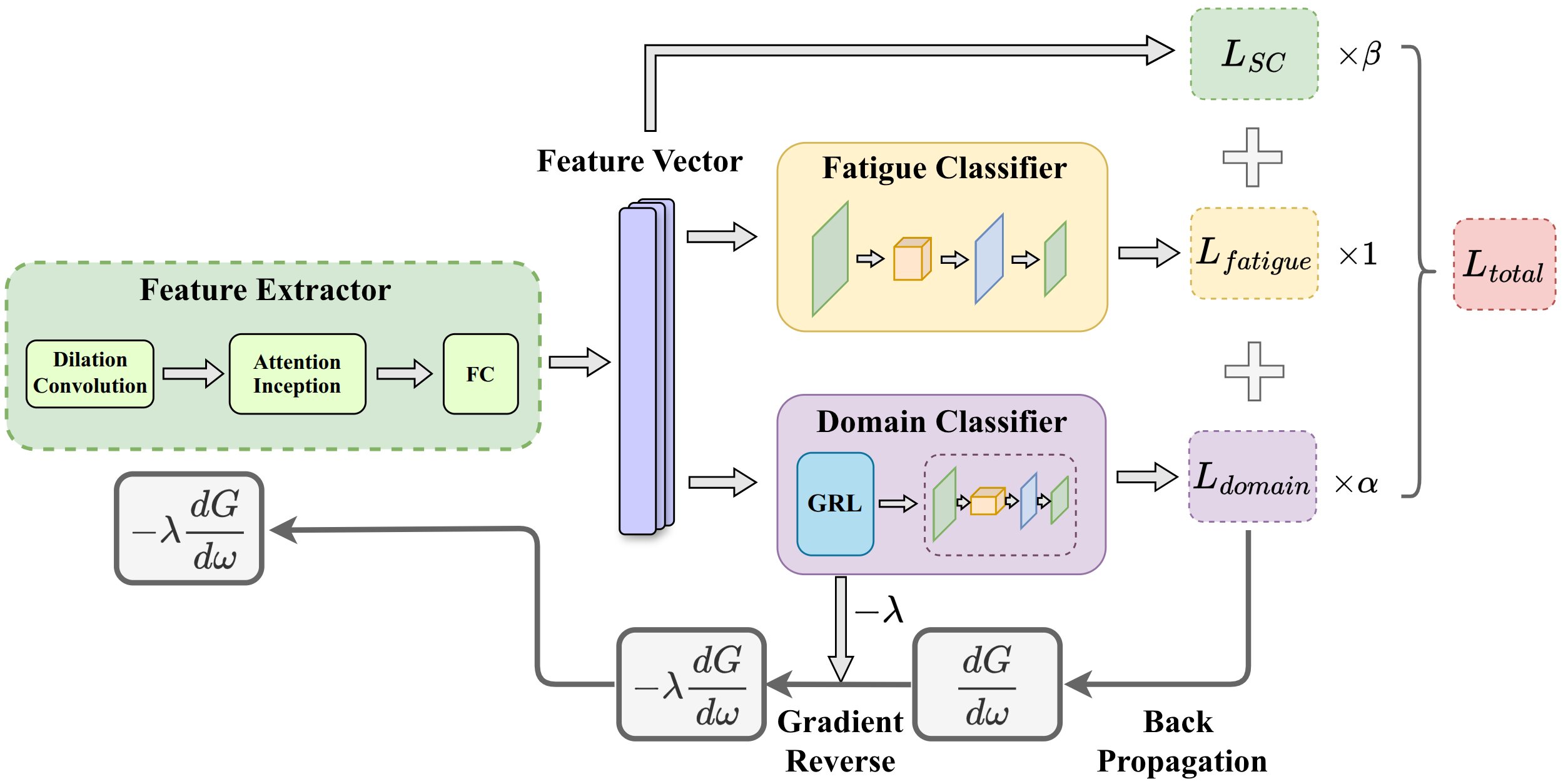}
    \caption{The adversarial and contrastive learning process.}
    \label{fig:training process}
\end{figure}

The proposed IADAN model is optimized using a joint loss function comprising two cross entropy (CE) losses and a SupCon loss, denoted as $L_{fatigue}$, $L_{domain}$ and $L_{SC}$. As illustrated in Fig.~\ref{fig:training process}, $L_{fatigue}$ performs fatigue detection tasks, while $L_{domain}$ eliminates subject-specific noise and $L_{SC}$ maximizes the similarity among positive samples and minimizes it for negative pairs, which further facilitates the extraction of generalized fatigue manifestation. The total loss is formulated as follows:
\begin{align}\label{eq:combined loss}
    L_{total} = L_{fatigue} + \alpha L_{SC} + \beta L_{domain}
\end{align}
where the hyperparameters $\alpha = 0.5$ and $ \beta = 0.8$ were empirically selected to achieve better performance.

\section{Experimental Setup}\label{sec:experiment}

To evaluate the performance of our proposed model, A fatigue experiment involving single-leg bodyweight calf raises was conducted. 12 healthy volunteers, 8 male and 4 female, were recruited for this study. The average age of the subjects was 19.83. The average heights were 176.40 cm for male and {164.25~cm} for female. The average weights were {65.50~kg} for males and {56.25~kg} for females. All subjects were free from lower-limb injuries or other underlying diseases and gave their informed consent for inclusion before they participated in the study. The protocol was approved by the Ethics Committee of of Wuhan University of Technology.

The experiment was conducted on a flat ground, with a chair positioned for the stability of the non-tested leg. sEMG and IMU signals were simultaneously recorded using Delsys Trigno wireless sensors at a sampling rate of 2000 Hz and LPMS-B2 IMU sensors at a sampling rate of 100 Hz. The subjective muscle fatigue degree was collected using the Borg CR-10 RPE scale.  Data preprocessing and model training were conducted on a computer equipped with NVIDIA GeForce RTX 3060 GPU. 

\begin{figure}[H]
    \centering
    \subfloat[]{\includegraphics[width=\linewidth]{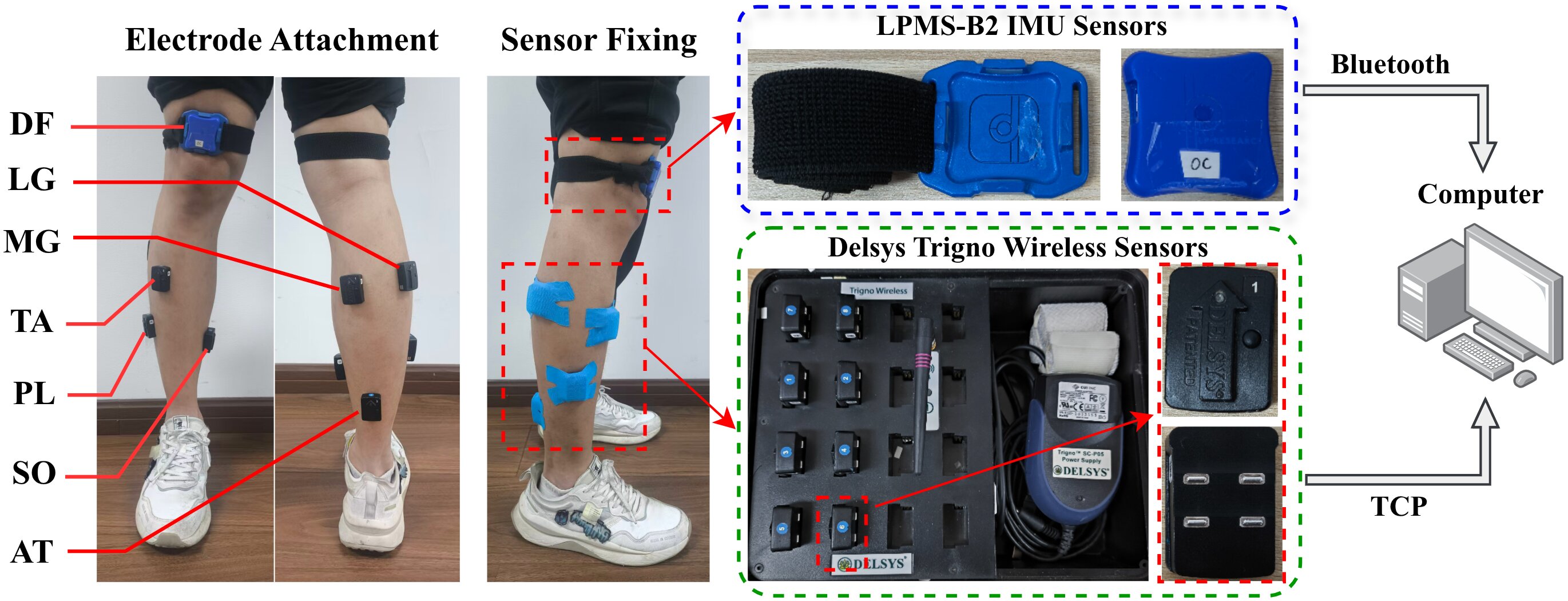}\label{fig:sensors}}
    \hfill
    \subfloat[]{\includegraphics[width=\linewidth]{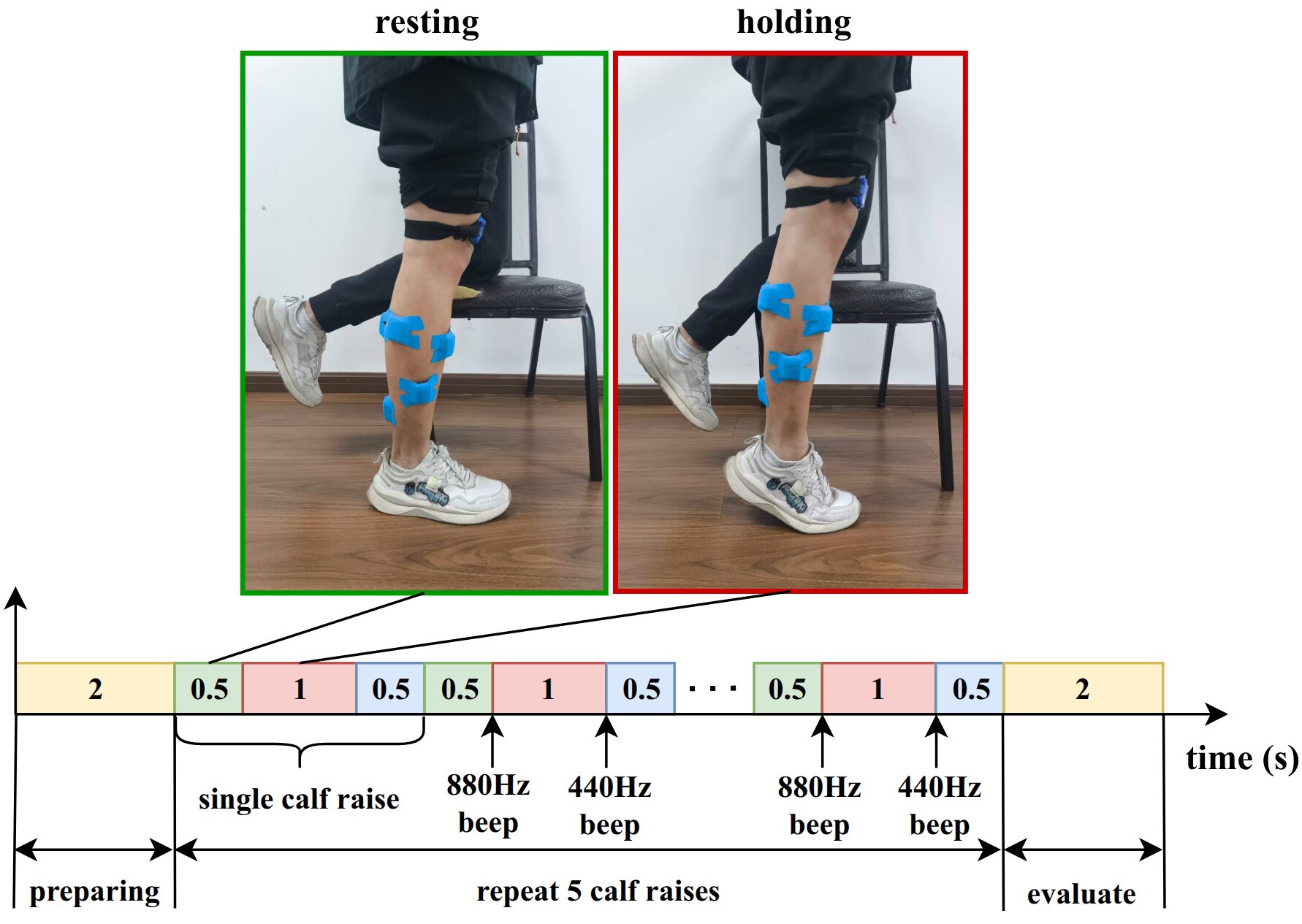}\label{fig:exp protocol}}
    \caption{The overall experimental setup. (a) Sensors attachment and data acquisition system. (b) Experiment protocol.}
    \label{fig:exp setup}
\end{figure}

The sEMG signals from six muscles of the lower limb were collected, including the medial gastrocnemius (MG), the lateral gastrocnemius (LG), the soleus (SO), the Achilles tendon (AT), the tibialis anterior (TA) and the peroneus longus (PL). To avoid noise interference, the skin was cleaned with 75\% alcohol wipes and body hair was removed prior to sensor electrodes attachment. All sEMG sensors were fixed with kinesiology tape to avoid offset. An IMU sensor was secured to the distal femur (DF) to monitor knee joint movement. Fig.~\ref{fig:sensors} shows the overall data acquisition system.

Initially, subjects performed several warm-up calf raises with their right leg to become familiar with the motion and were informed about the Borg CR-10 RPE scale, which ranges from 0, representing complete rest, to 10, denoting exhaustion, with specific description provided for each score \cite{frasie2024borg-cr10}. The entire process was structured into successive trials. Each trial consists of 2 seconds preparing, 5 repetitions of calf raise with each lasting 2 seconds, and 2 seconds evaluating, lasting a total of 14 seconds, as depicted in Fig.~\ref{fig:exp protocol}. In each calf raise process, subjects followed auditory prompts, including a high-pitched beep at 0.5s and a low-pitched beep at 1.5s. Subjects were required to perform calf raises and maintain raising status with their maximum effort once hearing high-pitched beeps, until they heard low-pitched beeps. The score of their current fatigue degree were collected after each 5 calf raises, regarded as the fatigue degree of this trial. Experiments started at degree 0 and ended when reaching degree 10 for each subject.

\begin{figure}[H]\label{fig:data preprocessing}
    \centering
    \subfloat[]{\includegraphics[width=\linewidth]{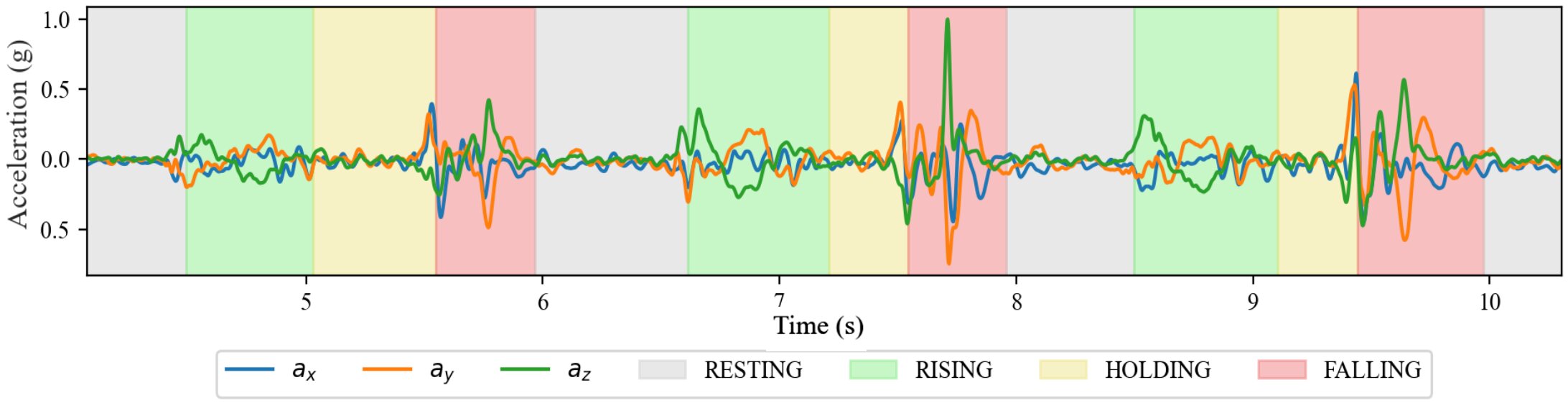}\label{fig:IMU segment}}
    \hfill
    \subfloat[]{\includegraphics[width=0.69\linewidth]{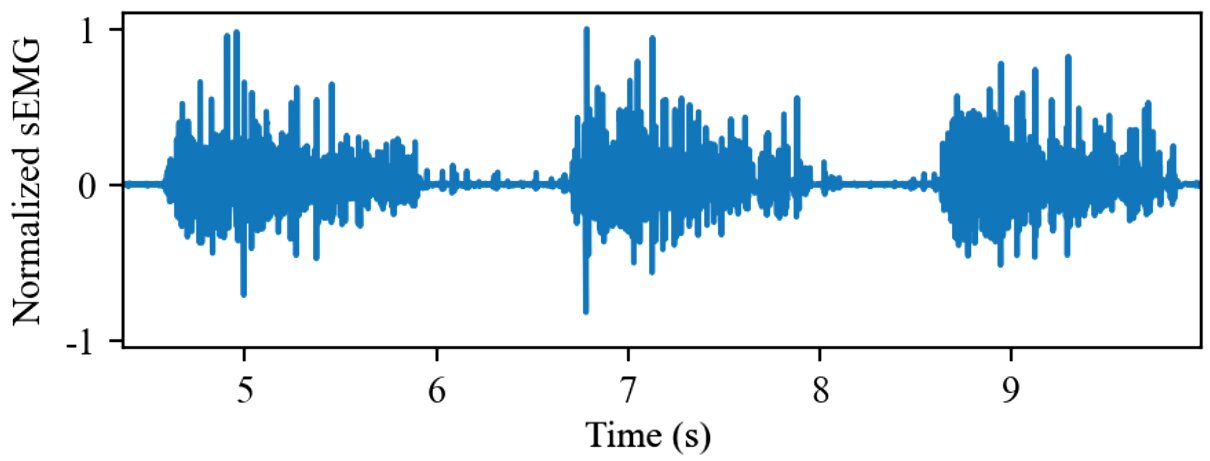}\label{fig:norm sEMG}}
    \hfill
    \subfloat[]{\includegraphics[width=0.29\linewidth]{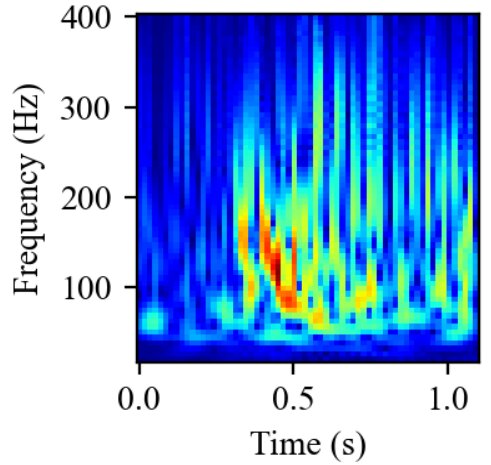}\label{fig:time-freq map}}
    \caption{The signal preprocessing protocol. (a) The segmented IMU data. (b) The normalized sEMG data (c) The time-frequency image.}
\end{figure}

the raw IMU signals were first processed with a 4th-order Butterworth lowpass filter with an 8 Hz cutoff frequency and were upsampled from 100 Hz to 500 Hz to temporal align with the sEMG signals. A state-machine was implemented to segment the IMU signals into four motion phases: resting, rising, holding, and falling. To visualize the segmentation results, the motion phases are shown as colored blocks illustrated in Fig.~\ref{fig:IMU segment}. The raw sEMG signals were processed with a 20--400 Hz 4th-order Butterworth bandpass filter and a 50 Hz notch filter to eliminate noise and power frequency interference. For each subject, root mean square (RMS) was calculated using a sliding window with a size of 100 samples and a stride of 50 samples for each channel, and the 95th percentile of the RMS across the entire process was estimated as the MVC used for normalization. Normalized signals were downsampled from 2000 Hz to 500 Hz, shown in Fig.~\ref{fig:norm sEMG}. Finally, the sEMG signals corresponding to holding phase were obtained and were performed continuous wavelet transform (CWT) to generate time-frequency images in Fig.~\ref{fig:time-freq map}.

CWT is a time-frequency analysis technique that decomposes a signal by convolving it with a set of localized oscillations known as wavelets. Different from the Fourier Transform, CWT captures when specific spectral components occur \cite{triwiyanto2017cwt}. The adoption of long wavelet windows for low-frequency components and short windows for high-frequency ones is controlled by parameter $a$ for scaling and $b$ for shifting, providing a balance between time and frequency resolution. The transformation is defined as:
\begin{align}
    W(a, b) = \frac{1}{\sqrt{|a|}} \int_{-\infty}^{\infty} x(t) \psi^* \left( \frac{t - b}{a} \right) dt
\end{align}
where $x(t)$ represents the input signals in the time domain and $\psi(t)$ denotes the mother wavelet. In this study, the Complex Morlet wavelet was selected, which works well for many biological signals \cite{triwiyanto2017cwt}, given by:
\begin{align}
    \psi(t) = \frac{1}{\sqrt{\pi B}} \cdot e^{j 2\pi C t} \cdot e^{-\frac{t^2}{B}}
\end{align}
where $B=1.5$ represents bandwidth parameter, and $C=1.0$ represents center frequency.

To minimize bias in subjective reporting, fatigue scores were categorized into three fatigue levels, and scores near the boundaries of each category were removed to ensure distinct categories. Ultimately, scores of [0,2] represent NF, scores of [4,6] represent MF, and scores of [8,10] represent SF. Domain labels were generated with each subject regarded as an independent domain. 
Before feeding into the network, the preprocessed time-frequency images were augmented through random shifting, scaling, and masking.
The total number of epochs was set to 100.

\section{Results}

\subsection{Model Performance Analysis}

\begin{figure}[H]
    \centering
    \subfloat[]{\includegraphics[width=0.46\linewidth]{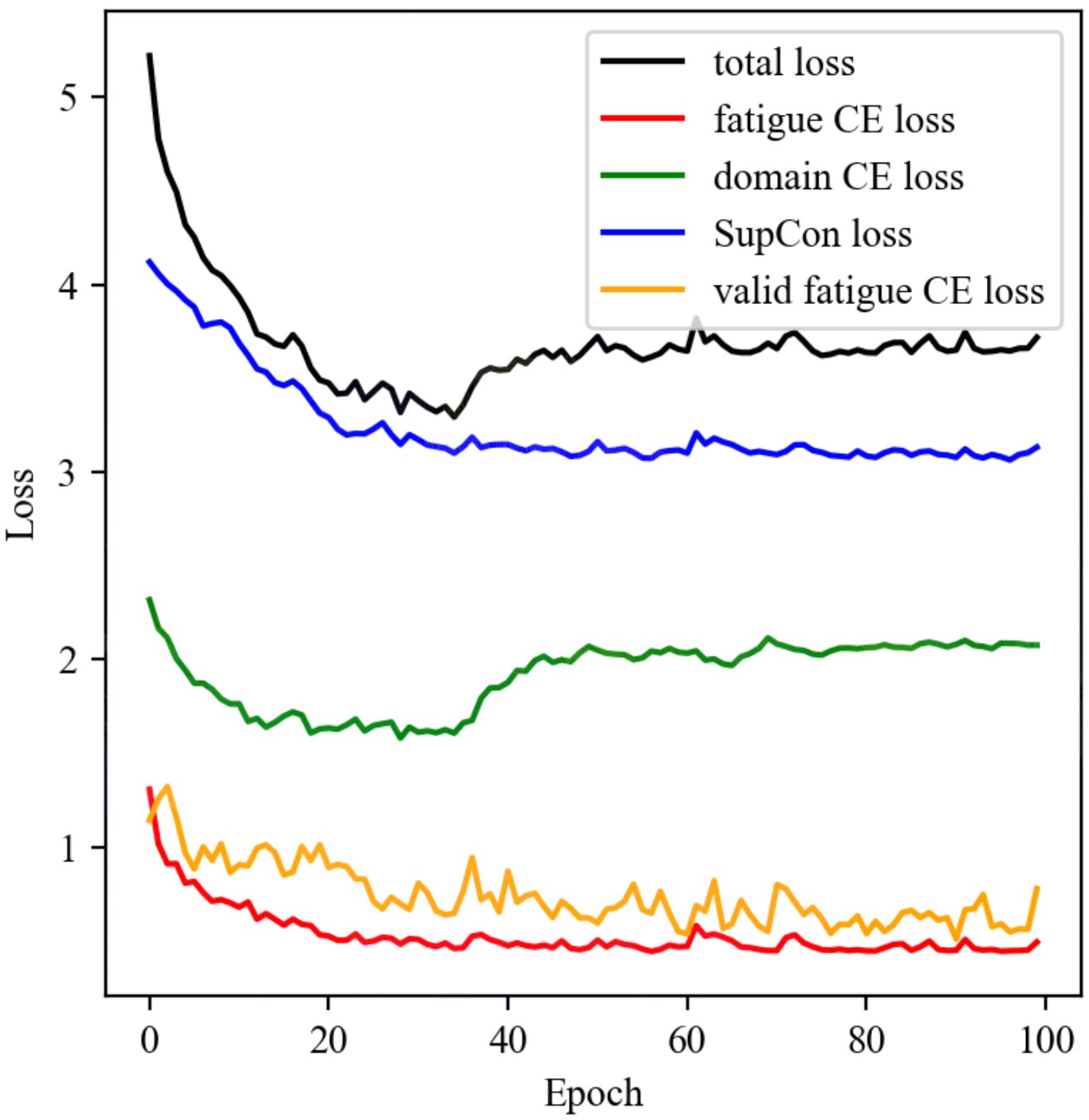}\label{fig:loss curve}}
    \subfloat[]{\includegraphics[width=0.47\linewidth]{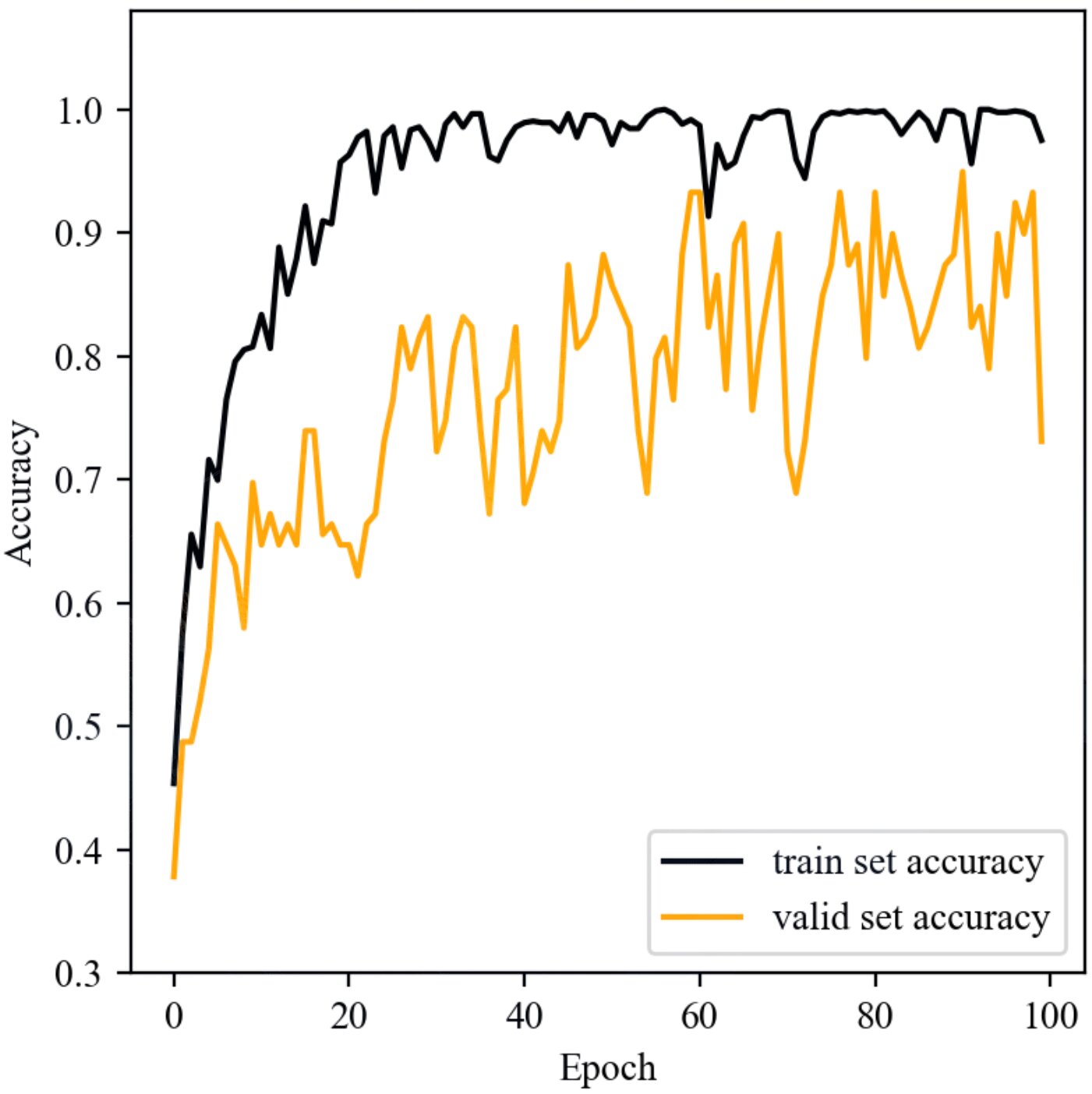}\label{fig:acc curve}}
    \caption{The Model performance analysis result. (a) The loss curve. (b) The accuracy curve.}
    \label{fig:loss&acc}
\end{figure}

To evaluate model performance, 4-fold cross-validation test was implemented. Each fold consisted of a training set of nine subjects and a validation set of the remaining three. The loss and accuracy curves are illustrated in Fig.~\ref{fig:loss&acc}. The detailed classification metrics, including accuracy, recall, and F1-score across the four experimental groups are summarized in Table~\ref{table:k-fold}.

\begin{table}[H]
    \begin{center}
        \caption{The Results of K-fold Cross Validation Experiments}\label{table:k-fold}
        \resizebox{0.93\linewidth}{!}{%
        \begin{tabular}{|c|c|c|c|}
            \hline
            \textbf{Experiment} &
             {\textbf{Accuracy(\%)}} &
             {\textbf{Recall(\%)}} &
             {\textbf{F1-score(\%)}} \\ \hline
            {Fold 1} & {95.28} & {94.91} & {95.09} \\ \hline
            {Fold 2} & {92.58} & {91.85} & {91.47} \\ \hline
            {Fold 3} & {92.64} & {91.95} & {92.20} \\ \hline
            {Fold 4} & {93.65} & {92.04} & {91.99} \\ \hline
             {\textbf{Average}} & \textbf{93.54}  & \textbf{92.69}  & \textbf{92.69}  \\ \hline
        \end{tabular}%
        }
    \end{center}
\end{table}

The result demonstrated that the proposed IADAN achieved an average accuracy of 93.54\%, with a recall of 92.69\% and an F1-score of 92.69\% in 3-class fatigue classification tasks. The fatigue CE loss and SupCon loss in Fig.~\ref{fig:loss curve} showed a downward trend as training progressed, while the domain CE loss initially declined but rapidly increased due to the adversarial training. The accuracy on the validation set in Fig.~\ref{fig:acc curve} increased steadily and finally converged to a high level, indicating good performance in cross-subject fatigue state recognition.

\begin{figure}[H]
    \subfloat[]{
        \includegraphics[width=0.47\linewidth]{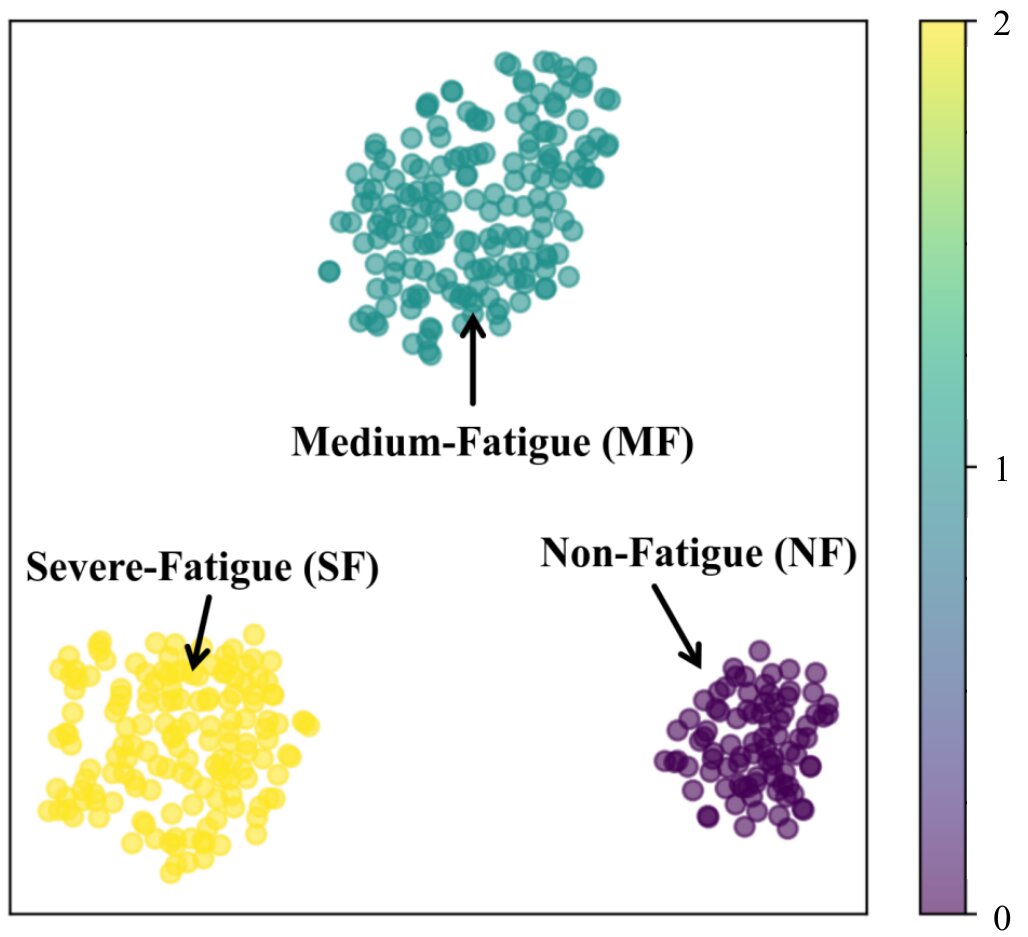}
        \label{fig:tsne_fatigue}}
    \hfill
    \subfloat[]{
        \includegraphics[width=0.47\linewidth]{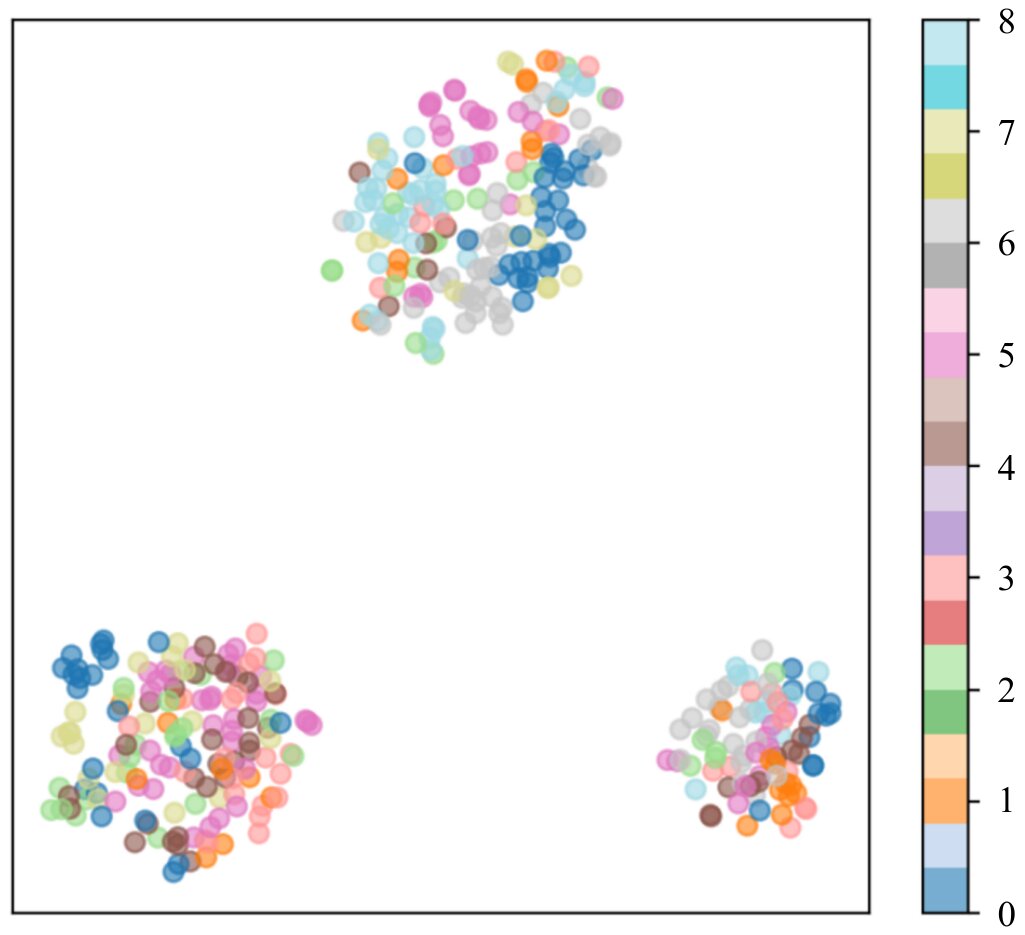}
        \label{fig:tsne_subject}
    }
    \caption{The t-SNE visualization result. (a) Feature vectors colored according to fatigue states. (b) Feature vectors colored according to subject ID.}
    \label{fig:t-SNE}
\end{figure}

To further analysis the generalization capability, t-distributed Stochastic Neighbor Embedding (t-SNE) are used to project high-dimensional feature vectors into 2D space for visualizing the distribution of feature vectors. Fig.~\ref{fig:tsne_fatigue} displays all projected vectors colored by fatigue state, while the same distribution in Fig.~\ref{fig:tsne_subject} are colored according to subject ID. The three fatigue states are distinctly separated into 3 clusters, whereas the subject classes are mixed together within each cluster, demonstrating the ability of our prposed model to capture subject-invariant fatigue features.

\subsection{Ablation Studies}

To further evaluate the contribution of each component in the proposed model, several ablation experiments were conducted, including structural ablation study and loss function ablation study. The results were calculated on the Fold-1 dataset.

In the structural ablation study, each structural component of IADAN was ablated to form several baseline models, including Inception-Attention-Net (IAN), Inception-Domain-Adversarial-Net (IDAN) and Attention-Domain-Adversarial-Net (ADAN), and these models were compared with the proposed IADAN. Experiment results are summarized in Table~\ref{table:structural ablation}. The results show that IADAN achieved the best performance across all metrics with an accuracy of 95.28\%, a recall of 94.91\% and an F1-score of 95.09\%. Without the attention module, the domain classifier or the Inception module, IDAN, IAN and ADAN all showed a significant performance drop in all metrics, particularly in IAN and ADAN with accuracies drops of 7.81\% and 15.34\% respectively, emphasizing the critical role of Inception-attention module and domain classifier when generalizing to unseen subjects.

\begin{table}[H]
    \begin{center}
        \caption{The Results of Structural Ablation Study}
        \label{table:structural ablation}
        \resizebox{0.93\linewidth}{!}{%
        \begin{tabular}{|c|c|c|c|}
            \hline
            \multicolumn{1}{|c|}{\textbf{Network}} & \multicolumn{1}{c|}{\textbf{Accuracy(\%)}} & \multicolumn{1}{c|}{\textbf{Recall(\%)}} & \multicolumn{1}{c|}{\textbf{F1-score(\%)}} \\ \hline
            IDAN  & 90.82  & 88.06  & 88.10  \\ \hline
            IAN   & 87.47  & 86.45  & 86.26  \\ \hline
            ADAN  & 79.94  & 73.95  & 75.94  \\ \hline
            \textbf{IADAN} & \textbf{95.28} & \textbf{94.91} & \textbf{95.09} \\ \hline
        \end{tabular}%
        }
    \end{center}
\end{table}

In our proposed training strategy, fatigue cross entropy (FCE) loss, domain cross entropy (DCE) loss and supervised contrastive (SC) loss are combined together to train the model, as illustrated in (\ref{eq:combined loss}). Thus, four configurations were compared: FCE, FCE+DCE, FCE+SC, and FCE+DCE+SC loss function. 
As shown in Table \ref{table: loss ablation}, model trained with the joint FCE+DCE+SC loss function outperformed all other configurations in accuracy, recall and F1-score, showing the improvement on generalization ability when domain cross entropy loss and supervised contrastive loss are employed. The two ablation studies above prove that IADAN has a superior performance in cross-subject fatigue detection tasks due to the proposed network structure and loss function, demonstrating the suitability and stability of our proposed network.

\begin{table}[H]
    \begin{center}
        \caption{The Results of Loss Function Ablation Study}
        \label{table: loss ablation}
        \resizebox{0.93\linewidth}{!}{%
        \begin{tabular}{|c|c|c|c|}
            \hline
            \textbf{Loss Function} &
             {\textbf{Accuracy(\%)}} &
             {\textbf{Recall(\%)}} &
             {\textbf{F1-score(\%)}} \\ \hline
              
            {FCE}     & {62.44} & {57.83} & {56.01} \\ \hline
            {FCE+DCE} & {81.87} & {82.12} & {81.47} \\ \hline
            {FCE+SC}  & {87.47} & {86.45} & {86.26} \\ \hline
            {\textbf{FCE+DCE+SC}} & {\textbf{95.28}} & {\textbf{94.91}} & {\textbf{95.09}} \\ \hline
        \end{tabular}%
        }
    \end{center}
\end{table}

\subsection{Comparison Experiments}

To evaluate the performance of our proposed model compared to other fatigue detection models, a comparative study was conducted including three architectures: ResNet augmented with a domain adversarial classifier, CLT-Net \cite{yu2024hybrid} and MFFNet \cite{zhang2021mffnet}. All models were trained and evaluated on Fold-1 dataset, summarized in Table \ref{table: model comparison}. 

\begin{table}[H]
    \begin{center}
        \caption{The Results of Model Comparison Experiments}
        \label{table: model comparison}
        \resizebox{0.93\linewidth}{!}{%
        \begin{tabular}{|c|c|c|c|}
            \hline
              
            { \textbf{Network}} & { \textbf{Accuracy(\%)}} & { \textbf{Recall(\%)}} & { \textbf{F1-score(\%)}} \\ \hline
              
            {ResNet}    & {86.32} & {84.03} & {83.85} \\ \hline
              
            {CLT-Net}   & {71.60} & {70.90} & {71.55} \\ \hline
              
            {MFFNet}    & {67.46} & {64.71} & {62.54} \\ \hline
             {\textbf{IADAN}} & \textbf{95.28}  & \textbf{94.91}  & \textbf{95.09}   \\ \hline
        \end{tabular}%
        }
    \end{center}
\end{table}

Compare to other fatigue detection models, the proposed IADAN consistently outperformed other models. Specifically, ResNet with domain adversarial classifier training results were 8.96\%, 10.87\% and 12.05\% lower than IADAN, indicating the multi-scale feature extraction ability of the proposed model. CLT-Net reached 71.60\% accuracy, 70.90\% recall, 71.55\% F1-score, respectively, while MFFNet even performed worse, which further demonstrates the strong capacity of IADAN in cross-subject fatigue recognition task, thereby providing the reliable guidance for practical rehabilitation training and assistance.

\section{Discussion \& Conclusion}

In this study, a novel neural network is proposed for robust cross-subject muscle fatigue detection, validated through a bodyweight calf raise experiment. By integrating dilated convolutional layers and an Inception-attention module, the model effectively extracts multi-scale features from sEMG signals. To overcome the challenge of inter-subject variability, a adversarial mechanism with a gradient reversal layer combined with supervised contrastive learning was implemented to extract subject-invariant features. 

The experimental results demonstrate the superior performance of our proposed model in 3-class fatigue classification tasks. Structural ablation and loss function ablation studies have proved the common feature extraction ability with accuracy exceeding all other ablated models. Without adversarial and supervised contrastive learning, the fatigue detection accuracy immediately drops even with strong feature extraction capability in time and frequency domains, demonstrating that the proposed model can recognozize cross-subject fatigue features through the adversarial training mechanism and supervised contrastive learning strategy, whereas to achieve robust fatigue detection, proven by accuracies drop of 23.68\%, 27.82\% in CLT-Net and MFFNet but only a 8.96\% drop in ResNet argumented with adversarial domain classifier.

Although IADAN has achieved a high accuracy in 3-class fatigue classification tasks, there are still some limitations. Adersatial learning via domain classifier often causes loss oscillations specifically when parameter $\lambda$ in GRL increases rapidly, 
even though it may bring about higher accuracy on cross-subject occasions. One way to solve this problem is to collect more data to stablize the training process. Another limitation is that more exertion patterns such as walking or jumping have not been explored due to experimental condition constraints such as inevitable sensors offset and skin sweating. Detecting fatigue states across both different subjects and different motions still remains a challenge to be addressed in future research.

\bibliographystyle{IEEEtran}
\bibliography{IEEEabrv, ref}

\begin{thebibliography}{10}
\providecommand{\url}[1]{#1}
\csname url@rmstyle\endcsname
\providecommand{\newblock}{\relax}
\providecommand{\bibinfo}[2]{#2}
\providecommand\BIBentrySTDinterwordspacing{\spaceskip=0pt\relax}
\providecommand\BIBentryALTinterwordstretchfactor{4}
\providecommand\BIBentryALTinterwordspacing{\spaceskip=\fontdimen2\font plus
\BIBentryALTinterwordstretchfactor\fontdimen3\font minus
  \fontdimen4\font\relax}
\providecommand\BIBforeignlanguage[2]{{%
\expandafter\ifx\csname l@#1\endcsname\relax
\typeout{** WARNING: IEEEtran.bst: No hyphenation pattern has been}%
\typeout{** loaded for the language `#1'. Using the pattern for}%
\typeout{** the default language instead.}%
\else
\language=\csname l@#1\endcsname
\fi
#2}}

\bibitem{constantin2021molecular}
D.~Constantin-Teodosiu and D.~Constantin, ``Molecular mechanisms of muscle
  fatigue,'' \emph{Int. J. Mol. Sci.}, vol.~22, no.~21, p. 11587, 2021.

\bibitem{frasie2024borg-cr10}
A.~Frasie, M.~Bertrand-Charette, M.~Compagnat, L.~J. Bouyer, and J.-S. Roy,
  ``Validation of the borg cr10 scale for the evaluation of shoulder perceived
  fatigue during work-related tasks,'' \emph{Appl. Ergon.}, vol. 116, p.
  104200, 2024.

\bibitem{zhang2024multilevel}
G.~Zhang, B.~Yang, P.~Zan, and D.~Zhang, ``Multilevel assessment of exercise
  fatigue utilizing multiple attention and convolution network {(MACNet)} based
  on surface electromyography,'' \emph{{IEEE} Trans. Neural Syst. Rehab. Eng.},
  vol.~33, pp. 243--254, 2024.

\bibitem{villafaina2023behavior}
S.~Villafaina, P.~Tomas-Carus, V.~Silva, A.~R. Costa, O.~Fernandes, and J.~A.
  Parraca, ``The behavior of muscle oxygen saturation, oxy and deoxy hemoglobin
  during a fatigue test in fibromyalgia,'' \emph{Biomedicines}, vol.~11, no.~1,
  p. 132, 2023.

\bibitem{Li2024non-invasive}
N.~Li, R.~Zhou, B.~Krishna, A.~Pradhan, H.~Lee, J.~He, and N.~Jiang,
  ``Non-invasive techniques for muscle fatigue monitoring: A comprehensive
  survey,'' \emph{ACM Comput. Surv.}, vol.~56, no.~9, p.~40, 2024.

\bibitem{Wu2025time_and_freq_features}
W.-T. Wu, S.-J. Ruan, Y.-W. Tu, Y.-L. Huang, and H.-J. Guo, ``A hybrid model
  for muscle fatigue detection: Integrating time and frequency domain features
  of semg signals with rpe scale during isometric contraction,'' \emph{{IEEE}
  Trans. Instrum. Meas.}, vol.~74, pp. 1--11, 2025.

\bibitem{sun2025detecting}
J.~Sun, C.~Zhang, G.~Liu, W.~Cui, Y.~Sun, and C.~Zhang, ``Detecting muscle
  fatigue during lower limb isometric contractions tasks: a machine learning
  approach,'' \emph{Front. Physiol.}, vol.~16, p. 1547257, 2025.

\bibitem{liu2023dynamic}
J.~Liu, Q.~Tao, and B.~Wu, ``Dynamic muscle fatigue state recognition based on
  deep learning fusion model,'' \emph{IEEE Access}, vol.~11, pp.
  95\,079--95\,091, 2023.

\bibitem{rampichini2020complexity}
S.~Rampichini, T.~M. Vieira, P.~Castiglioni, and G.~Merati, ``Complexity
  analysis of surface electromyography for assessing the myoelectric
  manifestation of muscle fatigue: A review,'' \emph{Entropy}, vol.~22, no.~5,
  p. 529, 2020.

\bibitem{zhang2021mffnet}
Y.~Zhang, S.~Chen, W.~Cao, P.~Guo, D.~Gao, M.~Wang, J.~Zhou, and T.~Wang,
  ``{MFFNet}: {Multi-dimensional} feature fusion network based on attention
  mechanism for {sEMG} analysis to detect muscle fatigue,'' \emph{Expert Syst.
  Appl.}, vol. 185, p. 115639, 2021.

\bibitem{yu2015dilation}
F.~Yu and V.~Koltun, ``Multi-scale context aggregation by dilated
  convolutions,'' \emph{arXiv preprint arXiv:1511.07122}, 2015.

\bibitem{woo2018cbam}
S.~Woo, J.~Park, J.-Y. Lee, and I.~S. Kweon, ``Cbam: Convolutional block
  attention module,'' in \emph{Proc. Eur. Conf. Comput. Vis.}, 2018, pp. 3--19.

\bibitem{szegedy2016inception}
C.~Szegedy, V.~Vanhoucke, S.~Ioffe, J.~Shlens, and Z.~Wojna, ``Rethinking the
  inception architecture for computer vision,'' in \emph{Proc. IEEE Conf.
  Comput. Vis. Pattern Recognit.}, 2016, pp. 2818--2826.

\bibitem{ganin2016domain}
Y.~Ganin, E.~Ustinova, H.~Ajakan, P.~Germain, H.~Larochelle, F.~Laviolette,
  M.~March, and V.~Lempitsky, ``Domain-adversarial training of neural
  networks,'' \emph{J. Mach. Learn. Res.}, vol.~17, no.~59, pp. 1--35, 2016.

\bibitem{khosla2020supervised}
P.~Khosla, P.~Teterwak, C.~Wang, A.~Sarna, Y.~Tian, P.~Isola, A.~Maschinot,
  C.~Liu, and D.~Krishnan, ``Supervised contrastive learning,'' \emph{Adv.
  Neural Inf. Process. Syst.}, vol.~33, pp. 18\,661--18\,673, 2020.

\bibitem{triwiyanto2017cwt}
T.~Triwiyanto, O.~Wahyunggoro, H.~A. Nugroho, and H.~Herianto, ``Continuous
  wavelet transform analysis of surface electromyography for muscle fatigue
  assessment on the elbow joint motion,'' \emph{Recent Adv. Electr. Electron.
  Eng.}, vol.~15, no.~3, p. 424, 2017.

\bibitem{yu2024hybrid}
J.~Yu, J.~Yang, C.~Zhu, W.~Meng, Q.~Liu, and Z.~Zhou, ``Hybrid
  {CNN-LSTM-Transformer} model for robust muscle fatigue detection during
  rehabilitation using {sEMG} signals,'' in \emph{Proc. 2024 Int. Conf. Adv.
  Robot. Mechatron. (ICARM)}, 2024, pp. 51--56.

\end{thebibliography}

\end{document}